\def\year{2018}\relax
\documentclass[letterpaper]{article} 
\usepackage{aaai18}  
\usepackage{times}  
\usepackage{helvet}  
\usepackage{courier}  
\usepackage{url}  
\usepackage{graphicx}  
\usepackage{subfigure}
\usepackage{algorithm}
\usepackage{algorithmic}

\usepackage{booktabs}

\usepackage{enumitem}
\usepackage{tikz}
\usetikzlibrary{fit,positioning}
\usepackage{amsmath}
\usepackage{amsfonts}
\usepackage{amssymb}
\usepackage{mathstyle}
\usepackage{mathtools}
\usepackage{cancel}
\usepackage{commath}
\usepackage{dsfont}
\usepackage{siunitx}
\usepackage{fancyhdr}
\usepackage{extramarks}
\usepackage{titlesec}
\usepackage{url}
\usepackage{algorithm}
\usepackage{algorithmic}
\usepackage{graphics}
\usepackage{framed}
\usepackage{chngcntr}
\usepackage{listings}
\usepackage{lipsum}
\usepackage{textcomp}

\usepackage{xcolor}
\usepackage{booktabs}
\usepackage{multirow}

\usepackage{wrapfig}
\usepackage[font=small,labelfont=bf]{caption} 

\usepackage{color}

\newcommand{\citep}{\cite}
\newcommand{\newcite}{\cite}
\newcommand{\citet}[1]{\citeauthor{#1} (\citeyear{#1})}

\begin{document}
%
\title{Knowledge-based Word Sense Disambiguation using Topic Models}
\author{Devendra Singh Chaplot, Ruslan Salakhutdinov\\
\{chaplot,rsalakhu\}@cs.cmu.edu\\
Machine Learning Department\\
School of Computer Science\\
Carnegie Mellon University}
\maketitle
\begin{abstract}
Word Sense Disambiguation is an open problem in Natural Language Processing which is particularly challenging and useful in the unsupervised setting where all the words in any given text need to be disambiguated without using any labeled data. Typically WSD systems use the sentence or a small window of words around the target word as the context for disambiguation because their computational complexity scales exponentially with the size of the context. In this paper, we leverage the formalism of topic model to design a WSD system that scales linearly with the number of words in the context. As a result, our system is able to utilize the whole document as the context for a word to be disambiguated. The proposed method is a variant of Latent Dirichlet Allocation in which the topic proportions for a document are replaced by synset proportions. We further utilize the information in the WordNet by assigning a non-uniform prior to synset distribution over words and a logistic-normal prior for document distribution over synsets. We evaluate the proposed method on Senseval-2, Senseval-3, SemEval-2007, SemEval-2013 and SemEval-2015 English All-Word WSD datasets and show that it outperforms the state-of-the-art unsupervised knowledge-based WSD system by a significant margin.
\end{abstract}

\section{Introduction}
\label{sec:Introduction}
Word Sense Disambiguation (WSD) is the task of mapping an ambiguous word in a
given context to its correct meaning.
WSD is an important problem in natural language processing (NLP), both in its
own right and as a stepping stone to more advanced tasks such as machine
translation \cite{Chan07wordsense}, information extraction and retrieval
\cite{Zhong:2012:WSD:2390524.2390563}, and question answering
\cite{Ramakrishnan03questionanswering}. WSD, being AI-complete~\cite{navigli2009word}, is still an open problem after over two decades of research. 
Following Navigli~\shortcite{navigli2009word}, we can roughly distinguish between supervised
and knowledge-based (unsupervised) approaches. Supervised methods require
sense-annotated training data and are suitable for lexical sample WSD tasks where systems are required to disambiguate a restricted set of target words. However, the performance of supervised systems is limited in the all-word WSD tasks as labeled data for the full lexicon is sparse and difficult to obtain. As the all-word WSD task is more challenging and has more practical applications, there has been significant interest in developing unsupervised knowledge-based systems. 
These systems only require an external knowledge source (such as WordNet) but no labeled training data. 

In this paper, we propose a novel knowledge-based WSD algorithm for the all-word WSD task, which utilizes the whole document as the context for a word, rather than just the current sentence used by most WSD systems. In order to model the whole document for WSD, we leverage the formalism of topic models, especially Latent Dirichlet Allocation (LDA). Our method is a variant of LDA in which the topic proportions for a document are
replaced by synset proportions for a document. We use a non-uniform prior for
the synset distribution over words to model the frequency of words within a synset.
Furthermore, we also model the relationships between synsets by using a
logistic-normal prior for drawing the synset proportions of the document. This
makes our model similar to the correlated topic model 
, with the difference that our priors are not
learned but fixed. In particular, the values of these priors are determined
using the knowledge from WordNet. We evaluate our system on a set of five benchmark datasets, Senseval-2, Senseval-3, SemEval-2007, SemEval-2013 and SenEval-2015 and show that the proposed model outperforms state-of-the-art knowledge-based WSD system.

\section{Related Work}
\label{sec:survey}
Lesk~\cite{lesk1986automatic} is a classical knowledge-based WSD algorithm which disambiguates a word by selecting a sense whose definition overlaps the most with the words in its context. Many subsequent knowledge-based systems are based on the Lesk algorithm. \citet{banerjee2003extended} extended Lesk by utilizing the definitions of words in the context and weighing the words by term frequency-inverse document frequency (tf-idf). \citet{basile2014enhanced} further enhanced Lesk by using word embeddings to calculate the similarity between sense definitions and words in the context.

The above methods only use the words in the context for disambiguating the target word. However, \citet{chaplot2015unsupervised} show that sense of a word depends on not just the words in the context but also on their senses. Since the senses of the words in the context are also unknown, they need to be optimized jointly. In the past decade, many graph-based unsupervised WSD methods have been developed which typically leverage the underlying structure of Lexical Knowledge Base such as WordNet and apply well-known graph-based techniques to efficiently select the best possible combination of senses in the context. \citet{navigli2007graph} and \citet{navigli2010experimental} build a subgraph of the entire lexicon containing vertices useful for disambiguation and then use graph connectivity measures to determine the most appropriate senses. \citet{mihalcea2005unsupervised} and \citet{sinha2007unsupervised} construct a sentence-wise graph, where, for each word every possible sense forms a vertex. Then graph-based iterative ranking and centrality algorithms are applied to find most probable sense. More recently, \citet{agirre2013random} presented an unsupervised WSD approach based on personalized page rank over the graphs generated using WordNet. The graph is created by adding content words to the WordNet graph and connecting them to the synsets in which they appear in
as strings. Then, the Personalized PageRank (PPR) algorithm is used to compute
relative weights of the synsets according to their relative structural
importance and consequently, for each content word, the synset with the highest PPR
weight is chosen as the correct sense. \citet{chaplot2015unsupervised} present a graph-based unsupervised WSD system which maximizes the total joint probability of all the senses in the context by modeling the WSD problem as a Markov Random Field constructed using the WordNet and a dependency parser and using a Maximum A Posteriori (MAP) Query for inference. Babelfy~\cite{moro2014entity} is another graph-based approach which unifies WSD and Entity Linking~\cite{rao2013entity}. It performs WSD by performing random walks with restart over BabelNet~\cite{NavigliPonzetto}, which is a semantic network integrating WordNet with various knowledge resources.

The WSD systems which try to jointly optimize the sense of all words in the context have a common limitation that their computational complexity scales exponentially with the number of content words in the context due to pairwise comparisons between the content words. Consequently, practical implementations of these methods either use approximate or sub-optimal algorithms or reduce the size of context. ~\citet{chaplot2015unsupervised} limit the context to a sentence and reduce the number of pairwise comparisons by using a dependency parser to extract important relations. \citet{agirre2013random} limit the size of context to a window of 20 words around the target word. \citet{moro2014entity} employ a densest subgraph heuristic for selecting high-coherence semantic interpretations of the input text. In contrast, the proposed approach scales linearly with the number of words in the context while optimizing sense of all the words in the context jointly. As a result, the whole document is utilized as the context for disambiguation. 

Our work is also related to \citet{boyd2007topic} who were the first to apply LDA techniques to WSD. In their approach, words senses that share
similar paths in the WordNet hierarchy are typically grouped in the same topic.
However, they observe that WordNet is perhaps not the most optimal structure
for WSD. Highly common, polysemous words such as ‘man’ and ‘time’ could
potentially be associated with many different topics making decipherment of
sense difficult. Even rare words that differ only subtly in their sense (e.g.,
quarterback -- the position and quarterback -- the player himself) could
potentially only share the root node in WordNet and hence never have a chance
of being on the same topic. 
\citet{cai2007improving} also employ the idea of global context for the task of WSD using topic models, but rather than using topic models in an unsupervised fashion, they embed topic features in a supervised WSD model.

\section{WordNet}
\label{sec:WordNet}

\begin{figure}
\begin{center}
    \includegraphics[width=1.0\linewidth]{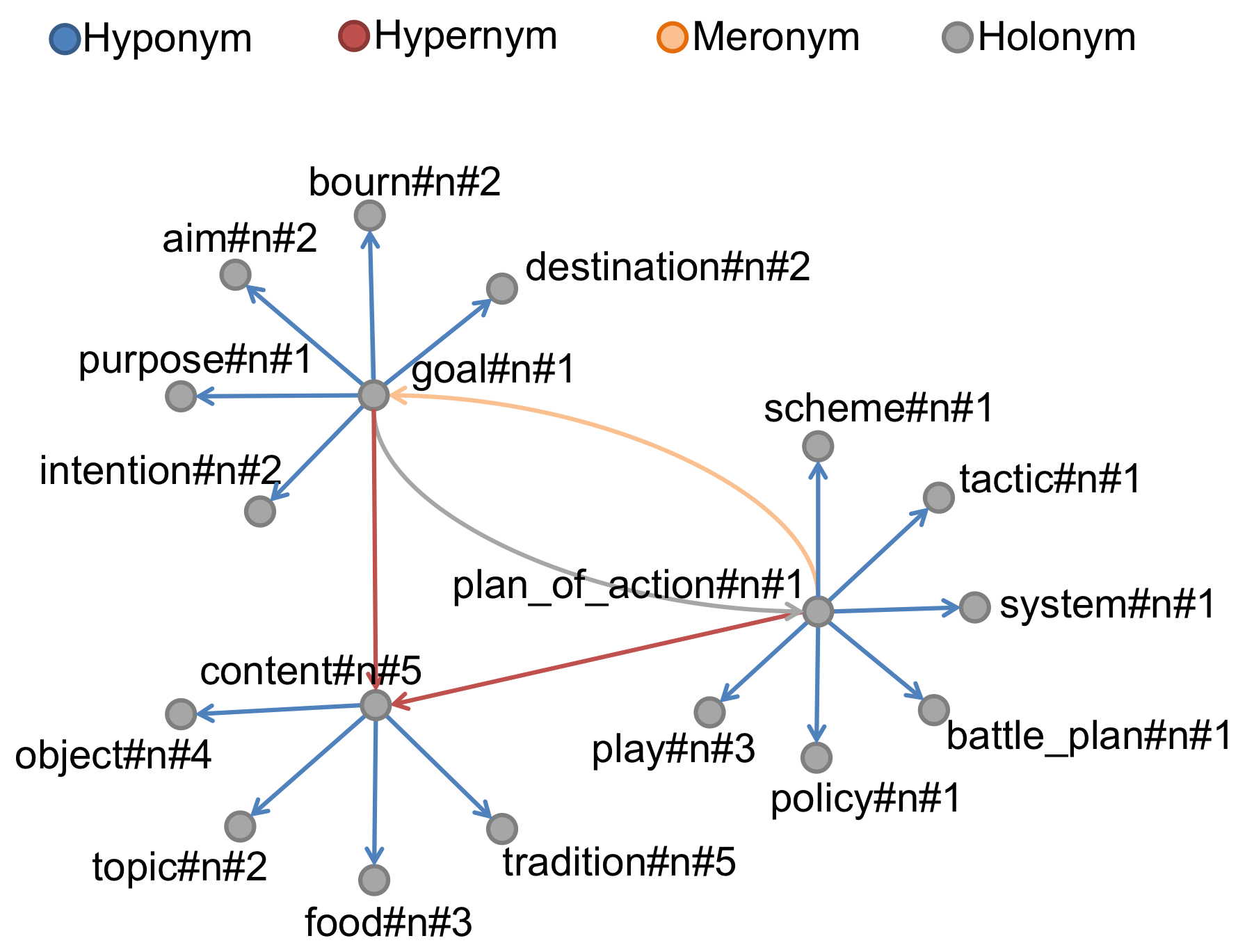}
\caption
{%
    WordNet example showing several synsets and the relations between them.%
}
\label{fig:wn}
\end{center}
\end{figure}

\begin{figure*}
\begin{center}
    \includegraphics[width=0.9\textwidth]{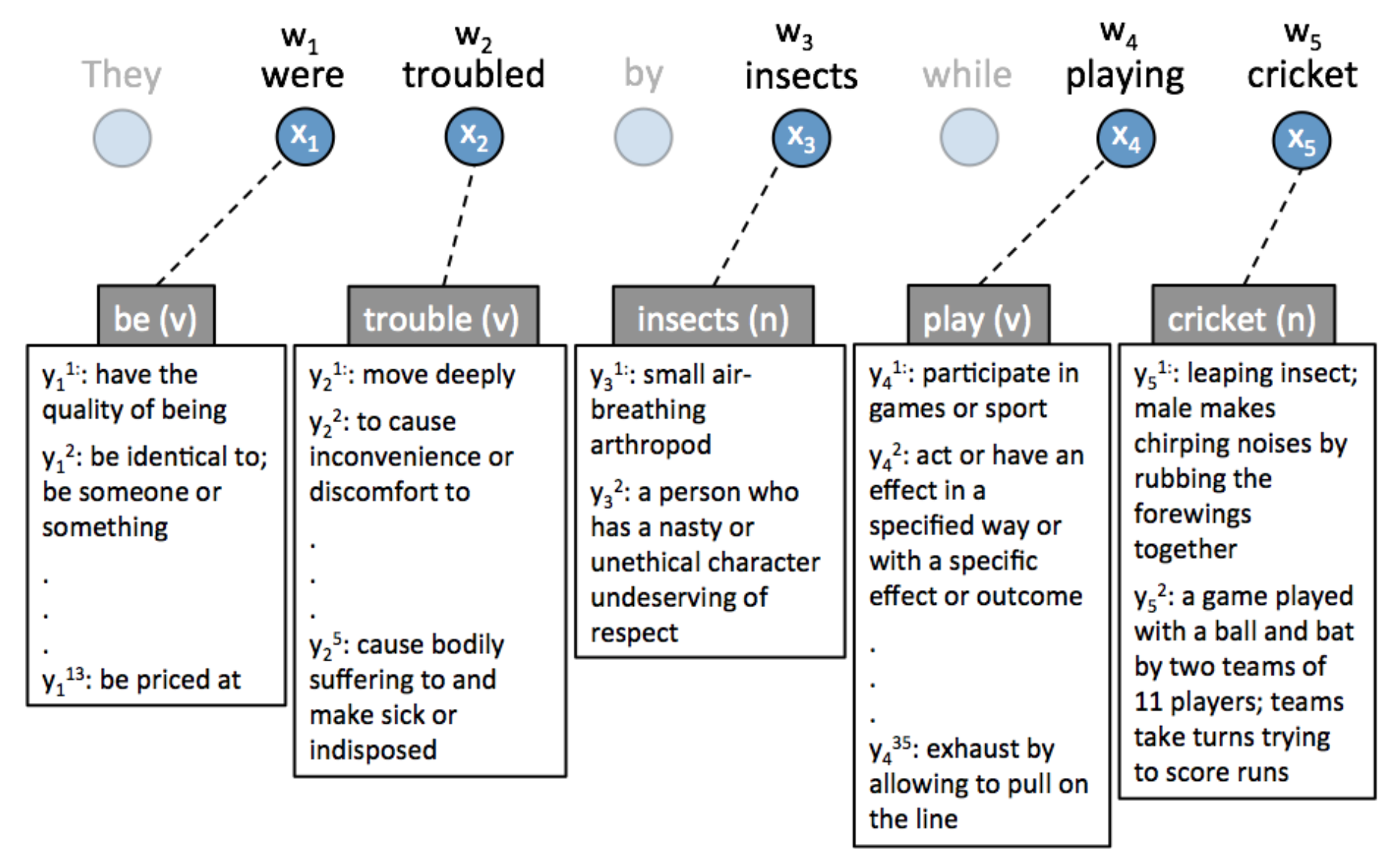}
    \caption
    {%
        An example of the all-word WSD task. Content words and their possible
        senses are labeled
        \(
            w
            _
            {
                i
            }
        \)
        and
        \(
            y
            _
            {
                i
            }
            ^
            {
                j
            }
        \),
        respectively.
    }
\label{fig:wsd_example}
\end{center}
\end{figure*}

Most WSD systems use a sense repository to obtain a set of possible senses for each word. WordNet is a comprehensive lexical database for the English language \cite{miller1995wordnet}, and is commonly used as the sense repository in WSD systems.

It provides a set of possible senses for each content word (nouns, verbs, adjectives and adverbs) in the language and classifies this set of senses by the POS tags. For example, the word ``cricket'' can have 2 possible noun senses: `cricket\#n\#1: leaping insect' and `cricket\#n\#2: a game played with a ball and bat', and a single possible verb sense, `cricket\#v\#1: (play cricket)'. Furthermore, WordNet also groups words which share the same sense into an entity called synset (set of synonyms). Each synset also contains a gloss and example usage of the words present in it. For example, `aim\#n\#2', `object\#n\#2, `objective\#n\#1', `target\#n\#5' share a common synset having gloss ``the goal intended to be attained''.

WordNet also contains information about different types of semantic relationships between synsets. These relations include hypernymy, meronymy, hyponymy, holonymy, etc. Figure~\ref{fig:wn} shows a graph of a subset of the WordNet where nodes denote the synset and edges denote different semantic relationship between synsets. For instance, `plan\_of\_action\#n\#1' is a meronym of `goal\#n\#1', and `purpose\#n\#1', `aim\#n\#1' and `destination\#n\#1' are hyponyms of `goal\#n\#1', as shown in the figure. These semantic relationships in the WordNet can be used to compute the similarity between different synsets using various standard relatedness measures \cite{pedersen2004wordnet}.

Note that although WordNet is the most widely used sense repository, the sense distinctions can be too fine-grained in many scenarios. This makes it difficult for expert annotators to agree on a correct sense, resulting in a very low inter-annotator agreement (~72\%) in standard WSD datasets. Nevertheless, we will use WordNet for our experiments for a fair comparison with previous work.   

\section{Methods}
\label{sec:Method}

\begin{figure*}
\begin{center}
\includegraphics[width=0.99\linewidth]{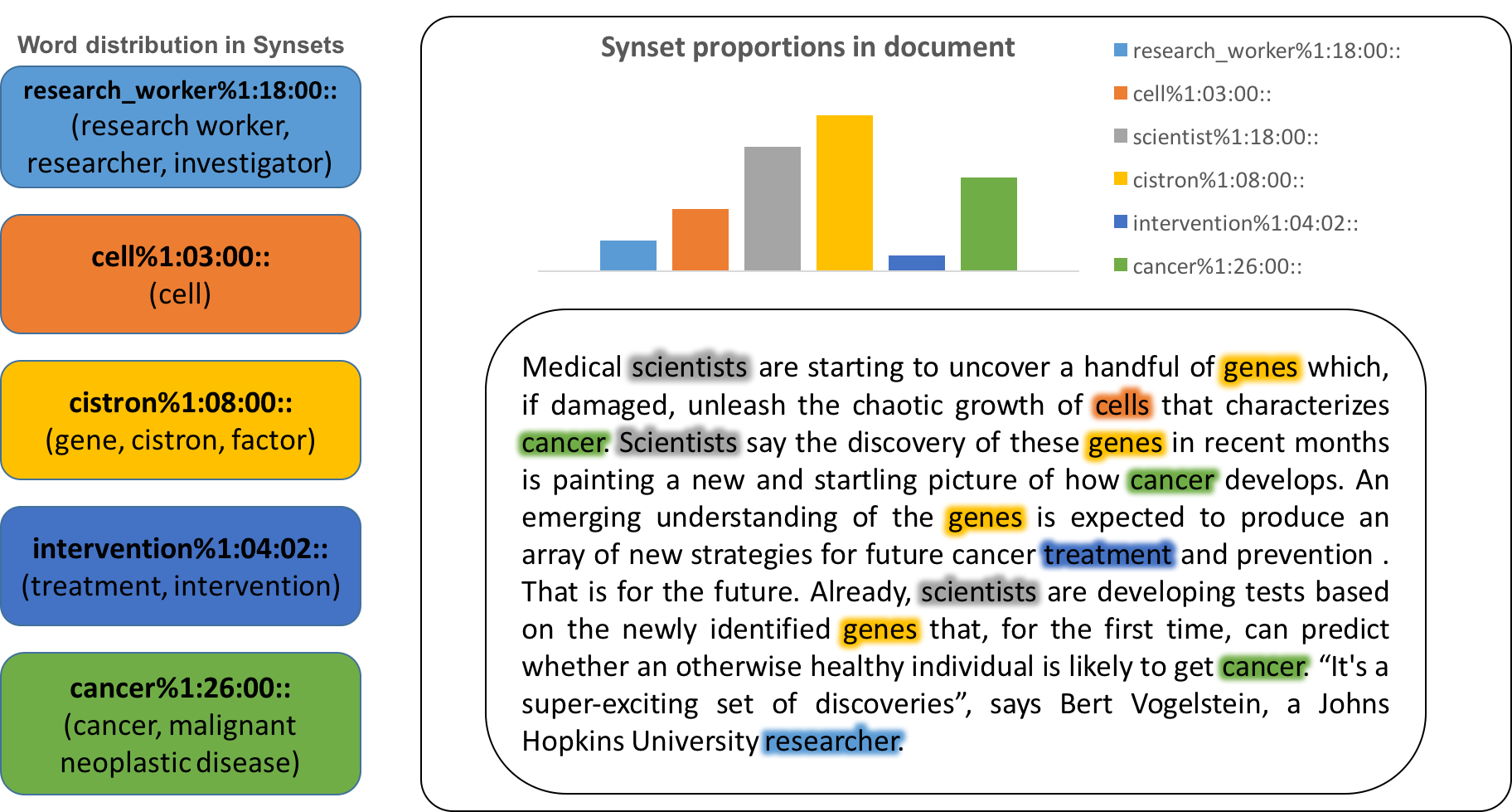}
\caption
{%
A toy example of word distribution in synsets and synset proportions in a document learned using the proposed model. Colors highlighting some of the words in the document denote the corresponding synsets they were sampled from.%
}
\label{fig:example}
\end{center}
\end{figure*}

\subsection{Problem Definition}
First, we formally define the task of all-word Word Sense Disambiguation by illustrating an example. Consider a sentence, where we want to disambiguate all the content words (nouns, verbs,
adjectives and adverbs):
\begin{quote}
    They were troubled by insects while playing cricket.
\end{quote}%
The sense
\(
    x
    _
    {
        i
    }
\)
of each content word (given its part-of-speech tag)
\(
    w
    _
    {
        i
    }
\)
can
take
\(
    k
    _
    {
        i
    }
\)
possible values from the set
\(
    \boldsymbol
    {y}_{i}=
    \{
        y
        _
        {
            i
        }
        ^
        {
            1
        }
        ,
        y
        _
        {
            i
        }
        ^
        {
            2
        }
        ,
        \ldots
        ,
        y
        _
        {
            i
        }
        ^
        {
            k_i
        }
    \}
\)
(see Figure~\ref{fig:wsd_example}).
In particular, the word
\(
    w
    _
    {
        5
    }
    =
    \text
    {%
        ``cricket''%
    }
\)
can either mean
\(
    y
    _
    {
        5
    }
    ^
    {
        1
    }
    =
\)
``a leaping insect''
or
\(
    y
    _
    {
        5
    }
    ^
    {
        2
    }
    =
\)
``a game played with a ball and bat played by two teams of 11 players.''
In this example, the second sense is more appropriate. The problem of mapping each content word in any given text to its correct sense is called the all-word WSD task. The set of possible senses for each word is given by a sense repository like WordNet.

\subsection{Semantics}
\label{sub:Methodology}
In this subsection, we describe the semantic ideas underlying the proposed method and how they are incorporated in the proposed model:
\begin{itemize}
    \item using the whole document as the context for WSD: modeled using Latent Dirichlet Allocation. 
    \item some words in each synset are more frequent than others: modeled using non-uniform priors for the synset distribution over words. 
    \item some synsets tend to co-occur more than others: modeled using logistic normal distribution for synset proportions in a document.
\end{itemize}
Wherever possible, we give examples to motivate the semantic ideas and illustrate their importance.

\subsubsection{Document context}
\label{sub:met1}
The sense of a word depends on other words in its context. In the WordNet, the
context of a word is defined to be the discourse that surrounds a language unit
and helps to determine its interpretation. It is very difficult to determine
the context of any given word. Most WSD systems use the sentence in which the
word occurs as its context and each sentence is considered independent of others. However, we know that a document or an article is about a particular topic in some domain and all the sentences in a document are not independent of each other. Apart from the words in the sentence, the occurrence of certain words in a document might help in word sense disambiguation. For example, consider the following sentence,
\begin{quote}
He forgot the chips at the counter.
\end{quote}

Here, the word `chips' could refer to potato chips, micro chips or poker chips. It is not possible to disambiguate this word without looking at other words in the document. The presence of other words like `casino', `gambler', etc. in the document would indicate the sense of poker chips, while words like `electronic' and `silicon' indicate the sense of micro chip. \citet{gale1992one} also observed that words strongly tend to exhibit only one sense in a given discourse or document. Thus, we hypothesize that the meaning of the word depends on words outside the sentence in which it occurs -- as a result, we use the whole document containing the word as its context.

\begin{figure*}
\centering
\begin{tikzpicture}
\tikzstyle{main}=[circle, minimum size = 10mm, thick, draw =black!100, node distance = 10mm]
\tikzstyle{connect}=[-latex, thick]
\tikzstyle{box}=[rectangle, draw=black!100]
  \node[main, draw=none] (alpha) [] {};
  \node[main, fill = black!10] (mu) [above of= alpha] {$\mu$};
  \node[main, fill = black!10] (sigma) [below of= alpha] {$\Sigma$};
  \node[main] (theta) [right=of alpha] {$\theta_m$};
  \node[main] (z) [right=of theta] {$z_{mn}$};
  \node[main, fill = black!10] (w) [right=of z] {$w_{mn}$};
  \node[main, draw=none] (gamma) [above of= w] {};
  \node[main] (beta) [above of= gamma] {$\beta_s$};
  \node[main, fill = black!10] (eta) [left=of beta] {$\eta_s$};
  \path (mu) edge [connect] (theta)
        (sigma) edge [connect] (theta)
        (theta) edge [connect] (z)
        (z) edge [connect] (w)
        (beta) edge [connect] (w)
        (eta) edge [connect] (beta);
  \node[rectangle, inner sep=0mm, fit= (beta) (eta),label=below right:S, xshift=12mm,  yshift=3mm] {};
  \node[rectangle, inner xsep=5mm, inner ysep=2.4mm, draw=black!100, fit= (beta) (eta), xshift=1mm] {};
  \node[rectangle, inner sep=0mm, fit= (z) (w),label=below right:$N_m$, xshift=9mm, yshift=3mm] {};
  \node[rectangle,  inner xsep=5mm, inner ysep=2.4mm, draw=black!100, fit= (z) (w),xshift=1mm] {};
  \node[rectangle, inner sep=2.6mm, fit= (z) (w),label=below right:M, xshift=13.5mm,  yshift=2mm] {};
  \node[rectangle, inner xsep=8.5mm, inner ysep=5mm, draw=black!100, fit = (theta) (z) (w),xshift=2.75mm] {};
\end{tikzpicture}
\caption
    {%
        Graphical model for the proposed method.
    }
    \label{fig:plate1}
\end{figure*}
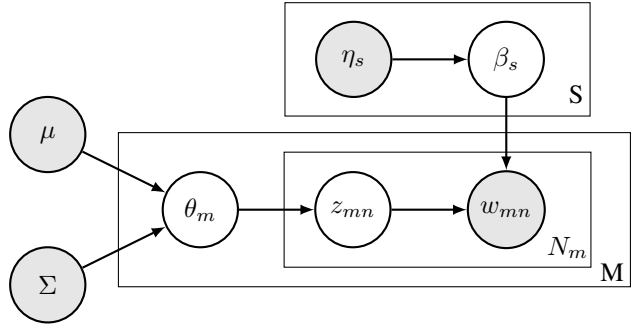

\subsubsection{Topic Models}
\label{sub:met2}
In order to use the whole document as the context for a word, we would like to model the concepts involved in the document. Topic models are suitable for this purpose, which aim to uncover the latent thematic structure in collections of documents \cite{blei2012probabilistic}.  The most basic example
of a topic model is Latent Dirichlet Allocation (LDA) \cite{blei2003latent}. It is based on the key assumption that
documents exhibit multiple topics (which are nothing but distributions over
some fixed vocabulary).

LDA has an implicit notion of word senses as words with several distinct meanings can appear in distinct topics (e.g., cricket the game in a ``sports'' topic and cricket the insect in a ``zoology'' topic). However, since the sense notion is only implicit (rather than a set of explicit senses for each word in WSD), it is not possible to directly apply LDA to the WSD task. Therefore, we modify the basic LDA by representing documents by synset probabilities rather than topic probabilities and consequently, the words are generated by synsets rather than topics. We further modify this graphical model to incorporate the information in the WordNet as described in the following subsections.

\subsubsection{Synset distribution over words}
\label{sub:met3}
Due to sparsity problems in large vocabulary size, the LDA model was extended
to a ``smoothed'' LDA model by placing an exchangeable Dirichlet prior on topic
distribution over words. In an exchangeable Dirichlet distribution, each
component of the parameter vector is equal to the same scalar. However, such a
uniform prior is not ideal for synset distribution over words since each synset
contains only a fixed set of words. For example, the synset defined as ``the
place designated as the end (as of a race or journey)'' contains only
`goal',`destination' and `finish'. Furthermore, some words in each synset are
more frequent than others. For example, in the synset defined as ``a person who
participates in or is skilled at some game'', the word `player' is more
frequent than word `participant', while in the synset defined as ``a theatrical
performer'', word `actor' is more frequent than word `player'. Thus, we decide
to have non-uniform priors for synset distribution over words.

\subsubsection{Document distribution over synsets}
\label{sub:met4}
The LDA model uses a Dirichlet distribution for the topic proportions of a
document. Under a Dirichlet, the components of the topic proportions vector are
approximately independent; this leads to the strong and unrealistic modeling
assumption that the presence of one topic is not correlated with the presence
of other topics. Similarly, in our case, the presence of one synset is correlated with the presence of others.
For example, the synset representing the `sloping land' sense of the word
`bank' is more likely to cooccur with the synset of `river' (a large natural
stream of water) than the synset representing `financial institution' sense of
the word `bank'. Hence, we model the correlations between synsets using a
logistic normal distribution for synset proportions in a document.

\subsection{Proposed Model}
\label{sub:Basic Model}
Following the ideas described in the previous subsection, we propose a probabilistic graphical model, which assumes that a corpus is generated according to the following process:
\begin{enumerate}
    \item For each synset, $s\ \in\ \{1,\ldots, S\}$
        \begin{enumerate}
            \item Draw word proportions $\beta_s \sim\ $Dir($\eta_s$)
        \end{enumerate}
    \item For each document, $m\ \in\ \{1,\ldots, M\}$
        \begin{enumerate}
            \item Draw $\alpha_m \sim\ \mathcal{N}(\mu,\Sigma)$
            \item Transform $\alpha_m$ to synset proportions $\theta_m = f(\alpha_m)$
            \item For each word in the document, $n\ \epsilon\ \{1,\ldots, N_m\}$
                \begin{enumerate}
                    \item Draw synset assignment $z_{mn} \sim\ $Mult($\theta_m$)
                    \item Draw word from assigned synset $w_{mn} \sim\ $Mult($\beta_{z_{mn}}$)
                \end{enumerate}
        \end{enumerate}
\end{enumerate}
where $f(\boldsymbol{\alpha}) =
\frac{\exp(\boldsymbol{\alpha})}{\sum_i\exp(\alpha_i)}$ is the softmax function.

Note that the prior for drawing word proportions for each sense is not
symmetric: $\eta_s$ is a vector of length equal to word vocabulary size, having
non-zero equal entries only for the words contained in synset $s$ in WordNet.
The graphical model corresponding to the generative process is shown in
Figure~\ref{fig:plate1}. Figure~\ref{fig:example} illustrates a toy example of a possible word distribution in synsets and synset proportions in a document learned using the proposed model. Colors highlighting some of the words in the document denote the corresponding synsets they were sampled from.

\subsection{Priors}
We utilize the information in the WordNet for deciding the priors for drawing
the word proportions for each synset and the synset proportions for each
document. The prior for distribution of synset $s$ over words is chosen as the
frequency of the words in the synset $s$, i.e.,
\begin{center}
$\eta_{sv}$ = Frequency of word $v$ in synset $s$.\\
\end{center}

The logistic normal distribution for drawing synset proportions has two priors,
$\mu$ and $\Sigma$. The parameter $\mu_s$ gives the probability of choosing a
synset $s$. The frequency of the synset $s$ would be the natural choice for
$\mu_s$ but since our method is unsupervised, we use a  uniform $\mu$ for all
synsets instead.
The $\Sigma$ parameter is used to model the relationship between synsets.
Since, the inverse of covariance matrix will be used in inference, we directly
choose $(i,j)$th element of inverse of covariance matrix as follows:
\begin{center}
$\Sigma_{ij}^{-1}$ = Negative of similarity between synset $i$ and synset $j$\\
\end{center}

The similarity between any two synsets in the WordNet can be calculated using a
variety of relatedness measures given in WordNet::Similarity library~
\cite{pedersen2004wordnet}. In this paper, we use the Lesk similarity measure as it is used in prior WSD systems. Lesk algorithm calculates the similarity between two synsets using the overlap between their definitions. 

\begin{table*}[]
\centering
\begin{tabular}{@{}clcccccl@{}}
\toprule
\multicolumn{1}{l}{}                                                       & \textbf{System} & \textbf{Senseval-2} & \textbf{Senseval-3} & \textbf{SemEval-07} & \textbf{SemEval-13} & \textbf{SemEval-15} & \textbf{All}                      \\ \midrule
\multirow{5}{*}{\begin{tabular}[c]{@{}c@{}}Knowledge\\ based\end{tabular}} & Banerjee03      & 50.6                & 44.5                & 32.0                & 53.6                & 51.0                & 48.7                              \\
                                                                           & Basile14        & 63.0                & 63.7                & \textbf{56.7}       & 66.2                & 64.6                & 63.7                              \\
                                                                           & Agirre14        & 60.6                & 54.1                & 42.0                & 59.0                & 61.2                & 57.5                              \\
                                                                           & Moro14          & 67.0                & 63.5                & 51.6                & \textbf{66.4}       & \textbf{70.3}       & 65.5                              \\ \cmidrule(l){2-8} 
                                                                           & WSD-TM          & \textbf{69.0}       & \textbf{66.9}       & 55.6                & 65.3                & 69.6                & \multicolumn{1}{c}{\textbf{66.9}} \\ \midrule
\multirow{3}{*}{Supervised}                                                & MFS             & 66.5                & 60.4                & 52.3                & 62.6                & 64.2                & \multicolumn{1}{c}{62.9}          \\
                                                                           & Zhong10         & 70.8                & 68.9                & 58.5                & 66.3                & 69.7                & \multicolumn{1}{c}{68.3}          \\
                                                                           & Melamud16       & 72.3                & 68.2                & 61.5                & 67.2                & 71.7                & \multicolumn{1}{c}{69.4}          \\ \bottomrule
\end{tabular}
\caption{Comparison of F1 scores with various WSD systems on English all-words datasets of Senseval-2, Senseval-3, SemEval-2007, SemEval-2013, SemEval-2015. WSD-TM corresponds to the proposed method. The best results in each column among knowledge-based systems are marked in \textbf{bold}.}
\label{table:results}
\end{table*}

\subsection{Inference}
We use a Gibbs Sampler for sampling latent synsets $z_{mn}$ given the values of
rest of the variables. Given a corpus of $M$ documents, the posterior over latent variables, i.e. the synset assignments $\boldsymbol{z}$, logistic normal parameter $\boldsymbol{\alpha}$, is as follows:
\begin{eqnarray*}
\begin{aligned}
& p(\boldsymbol{z}, \boldsymbol{\alpha}|\boldsymbol{w},\boldsymbol{\eta},\boldsymbol{\mu},\boldsymbol{\Sigma})\\
&\ \propto \quad p(\boldsymbol{w}|\boldsymbol{z},\boldsymbol{\beta})\;p(\boldsymbol{\beta}|\boldsymbol{\eta})\;p(\boldsymbol{z}|\boldsymbol{\alpha})\;p(\boldsymbol{\alpha}|\boldsymbol{\mu},\boldsymbol{\Sigma})\\
\end{aligned}
\end{eqnarray*}

\noindent The word distribution $p(w_{mn}|z_{mn},\beta)$ is multinomial in
$\beta_{z_{mn}}$ and the conjugate distribution $p(\beta_s|\eta_s)$ is
Dirichlet in $\eta_s$. Thus, $p(\boldsymbol{w}|\boldsymbol{z},\boldsymbol{\beta})\;p(\boldsymbol{\beta}|\boldsymbol{\eta})$ can be collapsed to $p(\boldsymbol{w}|\boldsymbol{z},\boldsymbol{\eta})$ by integrating out $\beta_s$ for all senses $s$ to obtain:
\begin{equation*}
p(\boldsymbol{w}|\boldsymbol{z},\boldsymbol{\eta}) = \prod_{s=1}^{S} \frac{\prod_{v} \Gamma(n_{sv}^{SV}+\eta_{sv})}{ \Gamma(n_{s}^{S}+||\eta_s||_1)}\frac{ \Gamma(||\eta_s||_1)}{\prod_{s} \Gamma(\eta_{sv})}
\end{equation*}

\noindent The sense distribution $p(z_{mn}|\boldsymbol{\alpha_m})$ is a multinomial distribution with
parameters $f(\boldsymbol{\alpha_m}) = \boldsymbol{\theta_m}$:
\begin{equation*}
p(\boldsymbol{z}|\boldsymbol{\alpha}) = \prod_{m=1}^{M}\Bigg(\prod_{n=1}^{N_m}\frac{exp(\alpha_m^{z_{mn}})}{\sum_{s=1}^S exp(\alpha_m^s)}\Bigg)
\end{equation*}

\noindent $\boldsymbol{\alpha_m}$ follows a normal distribution
which is not conjugate of the multinomial distribution: 
\begin{equation*}
p(\boldsymbol{\alpha_m}|\boldsymbol{\mu},\boldsymbol{\Sigma}) \sim\ \mathcal{N}(\boldsymbol{\alpha_m}|\boldsymbol{\mu},\boldsymbol{\Sigma})
\end{equation*}

\noindent Thus, $p(\boldsymbol{z}|\boldsymbol{\alpha})p(\boldsymbol{\alpha}|\boldsymbol{\mu},\boldsymbol{\Sigma})$ can't be collapsed. In typical logistic-normal topic models, a block-wise Gibbs sampling algorithm is used for
alternatively sampling topic assignments and logistic-normal parameters.
However, since in our case the logistic-normal priors are fixed, we can
sample synset assignments directly using the following equation:
\begin{eqnarray*}
\begin{aligned}
    & p(z_{mn} = k|rest) \\
&\ = \quad \frac{p(z,w|\alpha,\eta)}{p(z_{-mn},w|\alpha,\eta)}\\
&\ \propto \quad  p(z,w|\alpha,\eta)\\
&\ \propto \quad
    \frac{(\eta_{sv}+n_{sv_{-mn}}^{SV})}{n_{s_{-mn}}^{S}+||\eta_s||_1}\exp(\alpha_{m}^k)
\end{aligned}
\end{eqnarray*}

\noindent Here, $n_{sv}^{SV}$, $n_{sm}^{SM}$ and $n_{s}^{S}$ correspond to standard counts:
\begin{eqnarray*}
n_{sv}^{SV} & = &\sum_{m,n} \{z_{mn} = s, w_{mn} = v\}\\
n_{sm}^{SM} & = &\sum_{n} \{z_{mn} = s\}\\
n_{s}^{S} & = &\sum_{m} n_{sm}^{SM}
\end{eqnarray*}
\noindent The subscript $\cdot_{-mn}$ denotes the corresponding count without considering the word $n$ in document $m$. The value of $z_{mn}$ at the end of Gibbs sampling is the labelled sense of word $n$ in document $m$. The computational complexity of Gibbs sampling for this graphical model is linear with respect to number of words in the document \cite{qiu2014collapsed}.

\section{Experiments \& Results}
\label{sec:Results}

For evaluating our system, we use the English all-word WSD task benchmarks of
the SensEval-2 \cite{palmer-EtAl:2001:SENSEVAL}, SensEval-3
\cite{snyder-palmer:2004:Senseval-3}, SemEval-2007 \cite{pradhan2007semeval}, SemEval-2013 \cite{navigli2013semeval} and SemEval-2015 \cite{moro2015semeval}. \cite{raganato2017word} standardized all the above datasets into a unified format with gold standard keys in WordNet 3.0. We use the standardized version of all the datasets and use the same experimental setting as \cite{raganato2017word} for fair comparison with prior methods. 

In Table~\ref{table:results} we compare our overall F1 scores with different
unsupervised systems described in Section~\ref{sec:survey} which include Banerjee03~\cite{banerjee2003extended}, Basile14~\cite{basile2014enhanced}, Agirre14~\cite{agirre2013random} and Moro14~\cite{moro2014entity}. In addition to knowledge-based systems which do not require any labeled training corpora, we also report F1 scores of the state-of-the-art supervised systems trained on SemCor~\cite{miller1994using} and OMSTI~\cite{taghipour2015one} for comparison. Zhong10~\cite{zhong2010makes} use a Support Vector Machine over a set of features which include surrounding words in the context, their PoS tags, and local collocations. Melmaud16~\cite{melamud2016context2vec} learn context embeddings of a word and classify a test word instance with the sense of the training set word whose context embedding is the most similar to the context embedding of the test instance. We also provide the F1 scores of MFS baseline, i.e. labeling each word with its most frequent sense (MFS) in labeled datasets, SemCor~\cite{miller1994using} and OMSTI~\cite{taghipour2015one}. 

The proposed method, denoted by WSD-TM in the tables referring to WSD using topic models, outperforms the state-of-the-art WSD system by a significant margin (p-value \textless 0.01) by achieving an overall F1-score of 66.9 as compared to Moro14's score of 65.5. We also observe that the performance of the proposed model is not much worse than the best supervised system, Melamud16 (69.4). In Table~\ref{table:results1} we report the F1 scores on different parts of speech. The proposed system outperforms all previous knowledge-based systems over all parts of speech. This indicates that using document context helps in disambiguating words of all PoS tags.

\begin{table*}[]
\centering
\begin{tabular}{@{}clccccl@{}}
\toprule
\multicolumn{1}{l}{}                                                       & \textbf{System} & \textbf{Nouns} & \textbf{Verbs} & \textbf{Adjectives} & \textbf{Adverbs} & \textbf{All}                      \\ \midrule
\multirow{5}{*}{\begin{tabular}[c]{@{}c@{}}Knowledge\\ based\end{tabular}} & Banerjee03      & 54.1           & 27.9           & 54.6                & 60.3             & 48.7                              \\
                                                                           & Basile14        & 69.8           & 51.2           & 51.7                & 80.6             & 63.7                              \\
                                                                           & Agirre14        & 62.1           & 38.3           & 66.8                & 66.2             & 57.5                              \\
                                                                           & Moro14          & 68.6           & 49.9           & 73.2                & 79.8             & 65.5                              \\ \cmidrule(l){2-7} 
                                                                           & WSD-TM          & \textbf{69.7}  & \textbf{51.2}  & \textbf{76.0}       & \textbf{80.9}    & \multicolumn{1}{c}{\textbf{66.9}} \\ \midrule
\multirow{3}{*}{Supervised}                                                & MFS             & 65.8           & 45.9           & 72.7                & 80.5             & \multicolumn{1}{c}{62.9}          \\
                                                                           & Zhong10         & 71.0           & 53.3           & 77.1                & 82.7             & \multicolumn{1}{c}{68.3}          \\
                                                                           & Melamud16       & 71.7           & 55.8           & 77.2                & 82.7             & \multicolumn{1}{c}{69.4}          \\ \bottomrule
\end{tabular}
\caption{Comparison of F1 scores on different POS tags over all datasets. WSD-TM corresponds to the proposed method. The best results in each column among knowledge-based systems are marked in \textbf{bold}.}
\label{table:results1}
\end{table*}

\section{Discussions}
\label{sec:Discussions}

\begin{figure}
\begin{center}
    \includegraphics[width=\linewidth]{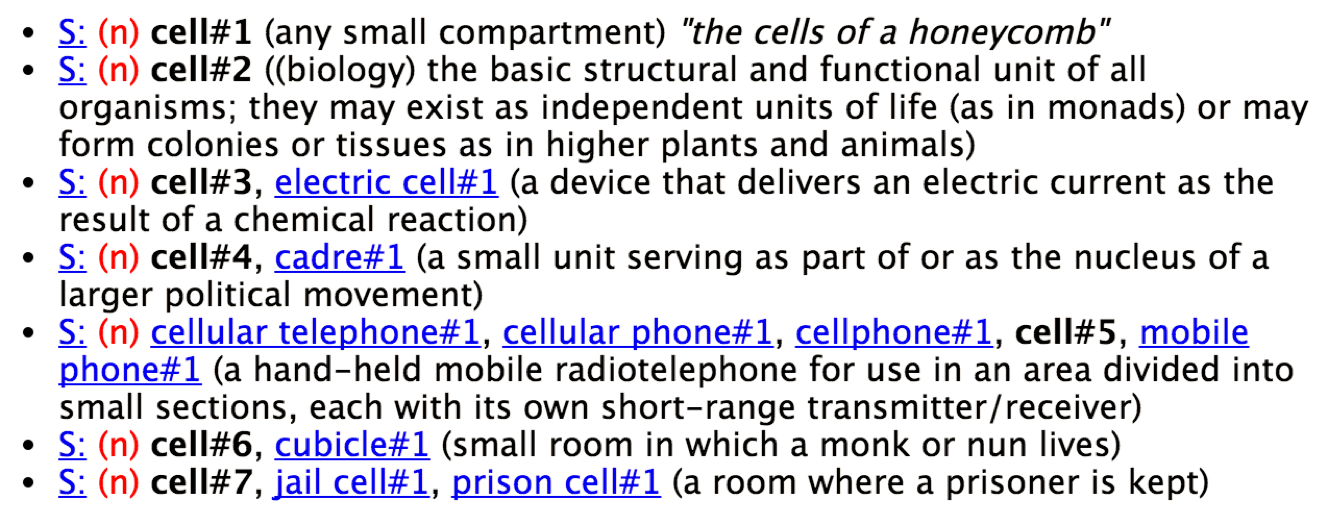}
\caption{The different senses of the word `cell' in the WordNet.}
\label{fig:cell}
\end{center}
\end{figure}

In this section, we illustrate the benefit of using the whole document as the context for disambiguation by illustrating an example. Consider an excerpt from the SensEval 2 dataset shown in Figure~\ref{fig:example}. Highlighted words clearly indicate that the domain of the document is Biology.  While most of these words are monosemous, let's consider disambiguating the word `cell', which is highly polysemous, having 7 possible senses as shown in Figure~\ref{fig:cell}. As shown in Table~\ref{tab:discussions}, the correct sense of cell (`cell\#2') has the highest
similarity with senses of three monosemous words `scientist', `researcher' and `protein'. The word `cell' occurs 21 times in the document, and several times, the other words in the sentence are not adequate to disambiguate it. Since our method uses the whole document as the context, words such as `scientists', `researchers' and `protein' help in disambiguating `cell', which is not possible otherwise.

The proposed model also overcomes several limitations of topic models based on Latent Dirichlet Allocation and its variants. Firstly, LDA requires the specification of the number of topics as a hyper-parameter which is difficult to tune. The proposed model doesn't require the total number of synsets to be specified as the total number of synsets are equal to the number of synsets in the sense repository which is fixed. Secondly, topics learned using LDA are often not meaningful as the words inside some topics are unrelated. However, synsets are always meaningful as they contain only synonymous words.
This is ensured in the proposed by using a non-uniform prior for word distribution in synsets.  

\begin{table}[]
\centering
\begin{tabular}{@{}lccc@{}}
\toprule
\multicolumn{1}{c}{\textbf{Sense of}} & \multicolumn{3}{c}{\textbf{Similarity with}}     \\ \cmidrule(l){2-4} 
\multicolumn{1}{c}{\textbf{`cell'}}   & scientist\#1   & researcher\#1  & protein\#1     \\ \midrule
cell\#1                               & 0.100          & 0.091          & 0.077          \\
cell\#2                               & \textbf{0.200} & \textbf{0.167} & \textbf{0.100} \\
cell\#3                               & 0.100          & 0.091          & 0.077          \\
cell\#4                               & 0.100          & 0.062          & 0.071          \\
cell\#5                               & 0.100          & 0.077          & 0.067          \\
cell\#6                               & 0.100          & 0.091          & 0.077          \\
cell\#7                               & 0.100          & 0.091          & 0.077          \\ \bottomrule
\end{tabular}
\caption{The similarity of different senses of the word `cell' with senses of three monosemous words `scientist', `researcher' and `protein'. The correct sense of cell, `cell\#2', has the highest similarity with all the three synsets.}
\label{tab:discussions}
\end{table}

\section{Conclusion}
\label{sec:Conclusion}
In this paper, we propose a novel knowledge-based WSD system based on a logistic normal topic model which incorporates semantic information about synsets as its priors. The proposed model scales linearly with the number of words in the context, which allows our system to use the whole document as the context for disambiguation and outperform state-of-the-art knowledge-based WSD system on a set of benchmark datasets. 

One possible avenue for future research is to use this model for supervised WSD. This could be done by using sense tags from the SemCor corpus as training data in a supervised topic model similar to the one presented by \newcite{mcauliffe2008supervised}. 
Another possibility would be to add another level to the hierarchy of the document generating process. This would allow us to bring back the notion of topics and then to define topic-specific sense distributions. 
The same model can also be extended to other problems such named-entity disambiguation.

\section{ Acknowledgments}
We would like to thank Prof. Eduard Hovy, Prof. Teruko Mitamura, Prof. Matthew Gormley, Zhengzhong Liu and Jakob Bauer for their extremely valuable comments and suggestions throughout the course of this project. 

{\small
	\bibliography{bibliography}
	\bibliographystyle{aaai}
} 
\end{document}